\documentclass[sigconf,nonacm]{acmart}
\usepackage{booktabs}
\usepackage{array}
\usepackage{xcolor}
\usepackage{tabularx}
\usepackage{longtable}
\usepackage{subcaption}
\usepackage{soul}
\usepackage[para]{footmisc}
\usepackage{cancel}
\usepackage[normalem]{ulem}
\usepackage{float}
\usepackage{seqsplit}

\makeatletter
\def\ps@plain{%
  \let\@mkboth\@gobbletwo
  \let\@oddhead\@empty
  \let\@evenhead\@empty
  \def\@oddfoot{\normalfont\hfil\thepage\hfil}%
  \let\@evenfoot\@oddfoot
}
\makeatother

\newcommand{\code}[1]{\texttt{\seqsplit{#1}}}

\newcommand{\etal}{\textit{et al.}}
\renewcommand\footnotetextcopyrightpermission[1]{}

\title{A Benchmark Framework for Screening Automation in Systematic Reviews}

\author{Gauransh Kumar}
\orcid{0009-0002-0232-9427}
\affiliation{%
  \institution{Université de Montréal}
  \city{Montreal}
  \country{Canada}
}
\email{gauransh.kumar@umontreal.ca}

\author{Luciano Marchezan}
\affiliation{%
  \institution{Université de Montréal}
  \city{Montreal}
  \country{Canada}
}
\email{lucianomarchp@gmail.com}

\author{Guillaume Genois}
\affiliation{%
  \institution{Université de Montréal}
  \city{Montreal}
  \country{Canada}
}
\email{guillaume.genois@umontreal.ca}

\author{Kévin Delcourt}
\affiliation{%
  \institution{Université de Montréal}
  \city{Montreal}
  \country{Canada}
}
\email{kevin.delcourt@umontreal.ca}

\author{Eugene Syriani}
\affiliation{%
  \institution{Université de Montréal}
  \city{Montreal}
  \country{Canada}
}
\email{syriani@iro.umontreal.ca}

\renewcommand{\shortauthors}{Kumar et al.}

\begin{abstract}
Systematic reviews (SR) are essential for evidence-based research, but their screening phase is highly time-consuming and labor-intensive. Large language models (LLMs) offer a promising opportunity to reduce this workload by assisting with article relevance classification. However, existing evaluation approaches often rely on traditional metrics that may be misleading for highly imbalanced SR screening datasets.This paper presents a benchmark dataset of $45\,064$ labeled entries for evaluating LLM performance in SR screening across 32 curated secondary studies. It proposes an evaluation framework that accounts for class imbalance, i.e., the natural prevalence of excluded articles relative to included articles in SRs. It also introduces PromptSR, a tool designed to support prompt experimentation, experiment management, and result analysis for LLM-based screening. We also present a use case demonstrating the application of SRBench and PromptSR.
\end{abstract}


\begin{document}
\thispagestyle{plain}
\maketitle

\section{Introduction}\label{sec:Introduction}
Systematic reviews (SRs), which include systematic literature reviews~\cite{Kitchenham2004}, systematic mapping studies~\cite{petersen2008systematic}, among others, are rigorous forms of literature review that follow structured methodologies to minimize bias and ensure reproducibility~\cite{Zhang2013}. They are widely used in software engineering~\cite{Kitchenham2009} to synthesize existing research and identify gaps. However, conducting an SR remains a time-consuming and labor-intensive process~\cite{borah2017analysis}. This is because the process involves multiple steps, including defining research questions, searching for relevant studies, screening studies based on predefined inclusion and exclusion criteria, extracting data from selected studies, and synthesizing the results.

Advances in natural language processing and machine learning, particularly through Large Language Models (LLMs), have shown promising results across a wide range of SR tasks~\cite{Syriani2024,Wang2023}. Researchers have demonstrated strong performance in text classification, summarization, and question-answering, often outperforming traditional machine learning approaches.
Given these capabilities, applying LLMs to automate key components of SRs presents a promising opportunity to improve both the efficiency and consistency of the review process.

Validating LLM performance in the context of SRs, however, requires appropriate benchmarks, including datasets and evaluation protocols that reflect the specific challenges of SR screening.
Toward this goal, Huotala~\etal~\cite{Huotala2025} introduced SESR--Eval (hereafter SESR), a benchmark dataset for evaluating LLMs in title-abstract screening for software engineering SRs, comprising $25\,683$ labeled primary studies from 18 SRs and enabling cross-study evaluation across multiple LLMs. Although it presents a useful dataset, SESR covers a restrictive subset of research domains in software engineering. In \cite{Syriani2024}, the authors have shown that LLM-based screening performance is highly dependent on the SR topic.
Additionally, the evaluation of SESR primarily relies on traditional classification metrics, such as accuracy, precision, recall, and F1-score. Since SR screening datasets are typically highly imbalanced~\cite{delaTorre2025}, with only a small proportion of relevant studies among thousands of candidates, these metrics may lead to misleading conclusions. In particular, accuracy can be artificially inflated when a model predominantly predicts the majority class of irrelevant studies~\cite{Madeyski2025}. While SESR represents an important contribution, there remains a need for benchmark studies that explicitly account for class imbalance and the specific priorities of SR screening. Furthermore, conducting such experiments with evolving LLMs and APIs introduces practical challenges related to experiment management, data storage, and result analysis. A dedicated tool is therefore needed to facilitate benchmark studies and make this line of research more accessible.


To address these limitations, we present a benchmark framework (\textit{\textbf{SRBench}}) for evaluating the performance of screening automation approaches (including LLMs) that accounts for the characteristics of highly imbalanced datasets. We incorporate metrics such as Matthews Correlation Coefficient (MCC)~\cite{Chicco2020} and Balanced Accuracy (BAcc), which better capture performance under class imbalance. The benchmark includes a dataset comprising $45\,064$ labeled primary studies from 32 secondary studies, extending the 18 SRs from SESR~\cite{Huotala2025} with 14 additional SRs to enable cross-study evaluation of LLM performance across a broader set of software engineering domains. We also introduce \textit{\textbf{PromptSR}}, a tool designed to support prompt experimentation, experiment management, and result analysis as part of SRBench. To validate the benchmark and its tool support, we present a use-case application of the framework using PromptSR, providing insights into how the framework can be used to evaluate LLM performance in SR screening.

The remainder of this paper is organized as follows. In Section~\ref{sec:Background}, we review background and related work on SRs, LLMs, and existing benchmarks. In Section~\ref{sec:slrbench}, we describe our benchmark dataset, including details on dataset collection and evaluation framework design. In Section~\ref{sec:promptslr}, we present the PromptSR tool. In Section~\ref{sec:results}, we report our use case application and discuss insights and limitations of SRBench. Finally, we conclude in Section~\ref{sec:Conclusion}.

\section{Background and Related Work}\label{sec:Background}

SRs are widely adopted in software engineering to synthesize existing evidence in a rigorous and reproducible manner, whether through in-depth analyses using systematic literature reviews~\cite{Kitchenham2009}, broad overviews using mapping studies~\cite{petersen2008systematic}, or other secondary studies. One critical phase of an SR is screening, in which researchers review titles and abstracts to select relevant primary studies from thousands of candidate papers retrieved through searches of digital libraries. This phase is typically framed as a binary classification task: relevant vs. irrelevant articles. However, screening is time-consuming and labor-intensive~\cite{Carver2013}, often requiring substantial manual effort from multiple reviewers. This creates risks of delays, human error, and reduced reproducibility, motivating research into automated or semi-automated screening approaches. For this reason, supporting tools have been developed to assist researchers during SRs~\cite{Marchezan2019,Bigendako2018}, including tools that leverage machine learning to prioritize studies for screening~\cite{Hamel2020,Chappell2023,delaTorre2025}.
Nevertheless, these tools often require substantial training data and may not generalize well across different SR topics. Recently, LLMs have been explored as supporting tools to assist in this phase~\cite{Scherbakov2025,Wang2023,Syriani2024}. While empirical studies report potential efficiency gains, their results also highlight limitations in terms of reliability, consistency, and generalizability across different SR topics~\cite{Lieberum2025}. Consequently, systematic and rigorous evaluation frameworks are needed to assess LLM performance in SR screening using appropriate datasets and metrics~\cite{Madeyski2025}. This motivates the need for a dedicated benchmark to guide future research in this area.

Existing SR tools address different aspects of the review process but leave a clear gap for structured LLM-based screening experiments. Process management tools such as ReLiS~\cite{Bigendako2018} and Covidence~\cite{covidence} support collaborative screening, protocol tracking, and data extraction, but are not designed for evaluating or comparing LLM configurations. AI-assisted tools such as Elicit~\cite{Whitfield2023} apply language models to literature search and abstract summarization, but focus on information retrieval rather than rigorous screening decisions, and lack mechanisms for domain-specific validation or controlled experimentation. Rayyan~\cite{Hamel2020} employs a machine learning classifier that learns from reviewer labels to prioritize studies, but does not support LLM backends or structured prompt ablation. None of these tools provides the infrastructure needed to systematically vary prompt configurations, execute experiments across multiple datasets, and evaluate results using metrics appropriate for imbalanced screening tasks. This gap motivates PromptSR, a tool designed to support reproducible prompt-engineering experiments in the context of LLM-based SR screening.

Several studies have investigated the role of LLMs in evidence synthesis workflows, including literature search, study selection, and data extraction~\cite{Rathi2024,Lieberum2025,Galli2025}. For instance, Lieberum~\etal~\cite{Lieberum2025} conducted a scoping review examining the application of generative models (predominantly GPT-based models) across multiple stages of medical SRs. Their findings suggest that although LLMs show promise, there is limited validation for large-scale applications, and their performance results are mixed. Similarly, Galli~\etal~\cite{Galli2025} analyze methodological challenges in deploying LLMs for title and abstract screening, emphasizing the influence of prompt engineering, validation protocols, model selection, and human-in-the-loop strategies on screening outcomes. While these studies report potential time savings, they also stress the importance of careful evaluation to avoid compromising review quality.

Considering SRs for software engineering, Huotala~\etal~\cite{Huotala2025} introduced SESR, a benchmark dataset designed to assess LLM performance for title-abstract screening.
SESR includes $25\,683$ labeled primary study entries derived from 18 secondary studies, enabling cross-study evaluation of nine LLMs. \footnote{Although they report $24$ studies comprising $34,528$ entries~\cite{Huotala2025}, their replication package does not provide the corresponding data~\cite{Huotala2025ZENODO}.}
Their results indicate that LLMs are relatively inexpensive to deploy and often exhibit comparable performance across models. However, screening effectiveness varies substantially across secondary studies, and none of the evaluated models achieve both consistently high recall and acceptable precision, suggesting that fully automated screening remains unreliable in practice. While SESR provides a valuable large-scale evaluation setting, its assessment is primarily based on standard classification metrics such as accuracy, precision, recall, and F$_1$, which biases results in imbalanced classifications.

These metrics are not fully aligned with the objectives of SR screening. In particular, they assume symmetric error costs and do not adequately capture the highly imbalanced nature of screening datasets, where relevant studies constitute only a small fraction of the candidate pool~\cite{Madeyski2025}. More critically, they fail to reflect the asymmetric consequences of screening errors, where false negatives can lead to the omission of relevant studies, while false positives primarily increase manual effort. This misalignment can lead to overly optimistic or misleading conclusions about model performance in practical review settings. Therefore, there is a need for evaluation frameworks that move beyond standard classification metrics and explicitly incorporate class imbalance and screening-specific error costs. In the following section, we present a benchmark designed to address these limitations while accounting for the practical requirements of LLM-assisted SR screening.

\section{SRBench}\label{sec:slrbench}

SRBench is a benchmark designed to provide a comprehensive evaluation framework for assessing the performance of automated screening tools, including LLMs, in the context of SRs. It comprises evaluation metrics and a dataset of labeled primary studies derived from published SRs in software engineering. Note that our benchmark also contains baseline results from a use case application, which are reported in Section~\ref{sec:results}.

\subsection{Evaluation Metrics}
\label{sec:metrics}
To overcome the limitations of standard metrics, we adopt a set of complementary metrics that together capture classifier behavior under class imbalance. These metrics include: \emph{Recall}, the ability to retrieve all relevant papers, which is critical in SR screening; \emph{Precision} and \emph{Negative Predictive Value} (NPV), which measure the reliability of positive and negative predictions respectively; \emph{Specificity}, which captures how well the model reduces unnecessary manual screening effort~\cite{Kusa2023}; \textit{BAcc}, the average of Recall and specificity, treating both classes equally regardless of their sizes and serving as the main measure of screening performance across datasets; and \textit{MCC} being the most informative metric for binary classification under class imbalance since it accounts for all four prediction outcomes proportionally to the actual class distribution~\cite{Madeyski2025}. By using BAcc and MCC as primary metrics, SRBench provides a more reliable and informative evaluation framework for assessing the performance of screening automation tools in highly imbalanced SR datasets. \emph{Accuracy} is also collected for completeness, but is not used as a primary indicator given its well-known sensitivity to class imbalance. With this, the benchmark provides a customizable evaluation framework that can be adapted to different research priorities and contexts, while ensuring the chosen metrics align with the specific challenges of SR screening.


\begin{table*}[t]
	\centering
	\caption{The SRBench dataset, including SESR and added SRs with size and class distribution.}
	\label{tab:taxonomy_clean}
	\resizebox{\textwidth}{!}{%
		\begin{tabular}{@{}lrrrr@{\hspace{0.8cm}}lrrrr@{}}
			\toprule
			\multicolumn{5}{c}{\textbf{SESR SR studies ($n=18$)}} &
			\multicolumn{5}{c}{\textbf{Addon SR studies($n=14$)}}                                                                                                                                               \\
			\cmidrule(r){1-5} \cmidrule(l){6-10}
			\textbf{Dataset}                                      & \textbf{Total} & \textbf{Include} & \textbf{Exclude} & \textbf{Inclusion Rate} &
			\textbf{Dataset}                                      & \textbf{Total} & \textbf{Include} & \textbf{Exclude} & \textbf{Inclusion Rate}                                                              \\
			\midrule
			\code{MLDefect}                                     & 1{,}194        & 742              & 452              & 0.621                   & \code{ARCHIML}         & 2{,}655 & 39  & 2{,}616 & 0.015 \\
			\code{SE4MVP}                                       & 222            & 190              & 32               & 0.856                   & \code{BEHAVE}          & 447     & 87  & 360     & 0.195 \\
			\code{DTCPS}                                        & 405            & 134              & 271              & 0.331                   & \code{CODECOMPR}       & 3{,}929 & 549 & 3{,}380 & 0.140 \\
			\code{ConflictModelMerge}                           & 322            & 171              & 151              & 0.531                   & \code{ESM\_2}          & 89      & 47  & 42      & 0.528 \\
			\code{CodeClone}                                    & 9{,}695        & 515              & 9{,}180          & 0.053                   & \code{ESPLE}           & 885     & 56  & 829     & 0.063 \\
			\code{CodeSmellDetect}                              & 1{,}651        & 167              & 1{,}484          & 0.101                   & \code{GAMESE}          & 345     & 31  & 314     & 0.090 \\
			\code{MaintainDT}                                   & 559            & 142              & 417              & 0.254                   & \code{LC}              & 2{,}682 & 150 & 2{,}532 & 0.056 \\
			\code{MDE4MobileRobot}                              & 2{,}260        & 69               & 2{,}191          & 0.031                   & \code{MODELGUIDANCE}   & 1{,}682 & 211 & 1{,}471 & 0.125 \\
			\code{DomainSpec}                                   & 321            & 20               & 301              & 0.062                   & \code{MODELINGASSIST}  & 2{,}100 & 129 & 1{,}971 & 0.061 \\
			\code{CertMLSafe}                                   & 1{,}024        & 188              & 836              & 0.184                   & \code{OODP}            & 624     & 29  & 595     & 0.046 \\
			\code{CodeCommQual}                                 & 2{,}079        & 71               & 2{,}008          & 0.034                   & \code{RL4SE}           & 1{,}105 & 94  & 1{,}011 & 0.085 \\
			\code{LearnSoftConfSpec}                            & 69             & 69               & 0                & 1.000                   & \code{SECSELFADAPT}    & 1{,}255 & 63  & 1{,}192 & 0.050 \\
			\code{MPM4CPS}                                      & 326            & 153              & 173              & 0.469                   & \code{TRUSTSE}         & 451     & 91  & 360     & 0.202 \\
			\code{SPL4IOT}                                      & 167            & 56               & 111              & 0.335                   & \code{UPDATECOLLABMDE} & 871     & 56  & 815     & 0.064 \\
			\code{TimePressSE}                                  & 4{,}186        & 211              & 3{,}975          & 0.050                   &                          &         &     &         &       \\
			\code{ModelAutoCode4Wireless}                       & 318            & 187              & 131              & 0.588                   &                          &         &     &         &       \\
			\code{OpinionMineSD}                                & 781            & 127              & 654              & 0.163                   &                          &         &     &         &       \\
			\code{TestNN}                                       & 104            & 104              & 0                & 1.000                   &                          &         &     &         &       \\
			\bottomrule
		\end{tabular}%
	}
\end{table*}
\subsection{Dataset}
Our dataset extends the one introduced by SESR~\cite{Huotala2025}, which compiled screening decisions and article metadata from 18 SRs. By adding 14 new SRs, the dataset now comprises 32 SRs covering a total of $45\,064$ primary studies. All articles in the SRBench include both a title and an abstract. The corpus size reported by Huotala~\etal~\cite{Huotala2025} exceeds that of their replication package~\cite{Huotala2025ZENODO} because entries lacking an abstract were excluded from the provided data.

\subsubsection{Dataset construction}
The 14 newly added SRs were identified by reviewing high-quality journal publications in software engineering (Journal of Systems and Software, Information and Software Technology, and Empirical Software Engineering), where data availability is presented as meta-information. From our search, we collected 21 SRs. We contacted the authors of each paper to obtain detailed information about the screening process, including the list of excluded articles (for each screening phase, if more than one), the reasons for exclusion/inclusion, conflict flags, and resolutions. We received 14 of 21 answers from the authors.

The heterogeneity of the collected SR datasets required a curation process to ensure data consistency and screening reliability. We established a semi-automated procedure to curate all the screening information from these 14 SRs. Where metadata was incomplete, we retrieved it from seven major scientific databases: IEEE Xplore, ACM Digital Library, ScienceDirect, Springer Link, Web of Science, Scopus, and PubMed Central. Metadata was standardized across all 32 datasets using a uniform schema, capturing title, abstract, keywords, authors, venue, year, DOI, references, and publisher information.

Construction involved a multi-step validation pass for each dataset: comparing extracted titles and author names against the original replication packages, resolving discrepancies, and manually recovering articles that the automated retrieval failed to locate. During this process, we identified and corrected errors in the original SESR dataset. \emph{Duplicate entries} were the most prevalent issue, particularly in larger SRs: \code{CODECOMPR} contained duplicates caused by Unicode and character-encoding normalization differences between sources, while \code{MODELGUIDANCE} and \code{ARCHIML} exhibited the same pattern. \emph{Non-article items} constituted a second recurring problem: \code{BEHAVE} and \code{MODELINGASSIST} included conference proceedings volumes, technical standards, and project websites, all listed alongside research articles in the original packages. \emph{Unfindable articles} formed a third category: \code{ARCHIML} provided only author last-name fragments for some entries, making retrieval impossible. Finally, \emph{non-English articles}, primarily from Chinese, German, and Portuguese sources, were found in \code{OODP}, \code{MODELINGASSIST}, and \code{ARCHIML} and excluded accordingly. Besides this, a few edge cases happened: the article count for \code{OODP} differs from the original publication because the dataset was reconstructed by re-running the search query at collection time; and \code{ESM\_2} contains only articles retrieved through snowballing, as the primary search results were unavailable in the replication package. After validation, the final dataset files were produced in a clean, standardized format.

\subsubsection{Data characteristics}

The additional SRs broaden the topical coverage of software engineering research beyond the original SESR dataset, which primarily focused on areas such as software testing, model-driven engineering, and program analysis. In contrast, the newly added SRs introduce 11 categories that were previously missing, including software architecture, behavior-driven development, low-code/no-code development, ethics in software engineering, and reinforcement learning for software engineering. Overall, the resulting benchmark spans 35 ACM CCS categories, providing substantially broader coverage of the software engineering research landscape. Table~\ref{tab:taxonomy_clean} summarizes the composition of the SRBench dataset,\footnote{More information is available in our replication package~\cite{10.5281/ZENODO.20434525} with references to all SRs} including both the original 18 SRs and the 14 added ones, describing their number of primary studies and inclusion rates. These numbers range from highly imbalanced datasets, such as \code{ARCHIML} (0.015) and \code{MDE4MobileRobot} (0.031), to highly inclusive datasets, such as \code{SE4MVP} (0.856) and \code{MLDefect} (0.621), in which all candidate studies were retained. This data diversity enables a comprehensive evaluation of screening automation tools across diverse review contexts and inclusion rates.

\section{The PromptSR Tool}\label{sec:promptslr}



\begin{figure*}[h]
	\centering
	\includegraphics[width=.85\textwidth]{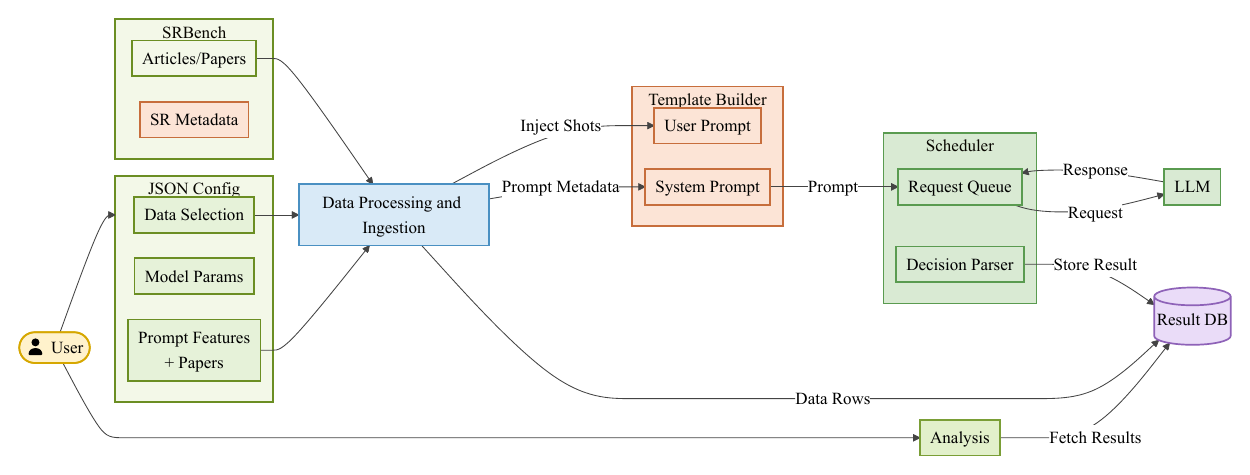}
	\caption{Technical Architecture of PromptSR}
	\label{fig:promptslr-arch}
\end{figure*}

Running experiments on SRBench requires orchestrating prompt configurations, executing screening tasks across 32 datasets, and computing SR-specific metrics. PromptSR is a tool designed to support this workflow, covering experiment configuration, execution, and result analysis without requiring custom scripting.  The tool follows a modular architecture (Figure~\ref{fig:promptslr-arch}) and is implemented primarily in Python. A data processing module handles dataset cleaning, formatting, and shot sampling for few-shot configurations. A template engine combines the selected feature configurations with a base prompt template to produce the final prompt submitted to the model, including role separation for multi-role interactions. A scheduler module manages asynchronous experiment execution, error detection, automatic retries, and incremental result storage. An analysis and visualization engine aggregates results and exposes the full suite of metrics and plots through the web interface. Data persistence is handled by PostgreSQL, with database access managed via Prisma, and the web interface is built with Streamlit. The complete source code is publicly available via Zenodo~\cite{10.5281/ZENODO.20434525}.

PromptSR supports a rich set of configurable prompt features that researchers can combine to construct and compare prompt variants. Contextual features include a \textit{Context} component, which assigns a persona and situates the task within the SR topic, a \textit{Description} that provides domain background, and \textit{Research Questions} that guide the classification decision, optionally accompanied by a short or long research question description for varying levels of domain elaboration. Criteria features allow the prompt to carry \textit{Inclusion Criteria}, \textit{Exclusion Criteria}, or both, expressing the rules that determine article relevance. Few-shot features control whether the prompt includes \textit{Positive Shots} (example included articles), \textit{Negative Shots} (example excluded articles), or a balanced combination of both. Output and reasoning features shape how the model responds: \textit{Chain of Thought} guides the model through step-by-step screening questions before committing to a decision, \textit{Uncertainty Output} allows borderline answers such as \code{MAYBE\_INCLUDE} or \code{MAYBE\_EXCLUDE}, \textit{Confidence Score} requests a numeric score alongside the decision, \textit{JSON Output Format} enforces structured parseable responses, and \textit{Leniency} instructs the model to err toward inclusion when uncertain. While these features are grounded in established SR guidelines~\cite{Kitchenham2004,petersen2008systematic}, the set is not fixed: new features can be introduced by modifying the prompt template and providing appropriate configuration files.

\begin{figure*}[t]
	\centering
	\begin{subfigure}[t]{.75\linewidth}
		\centering
		\includegraphics[width=\linewidth]{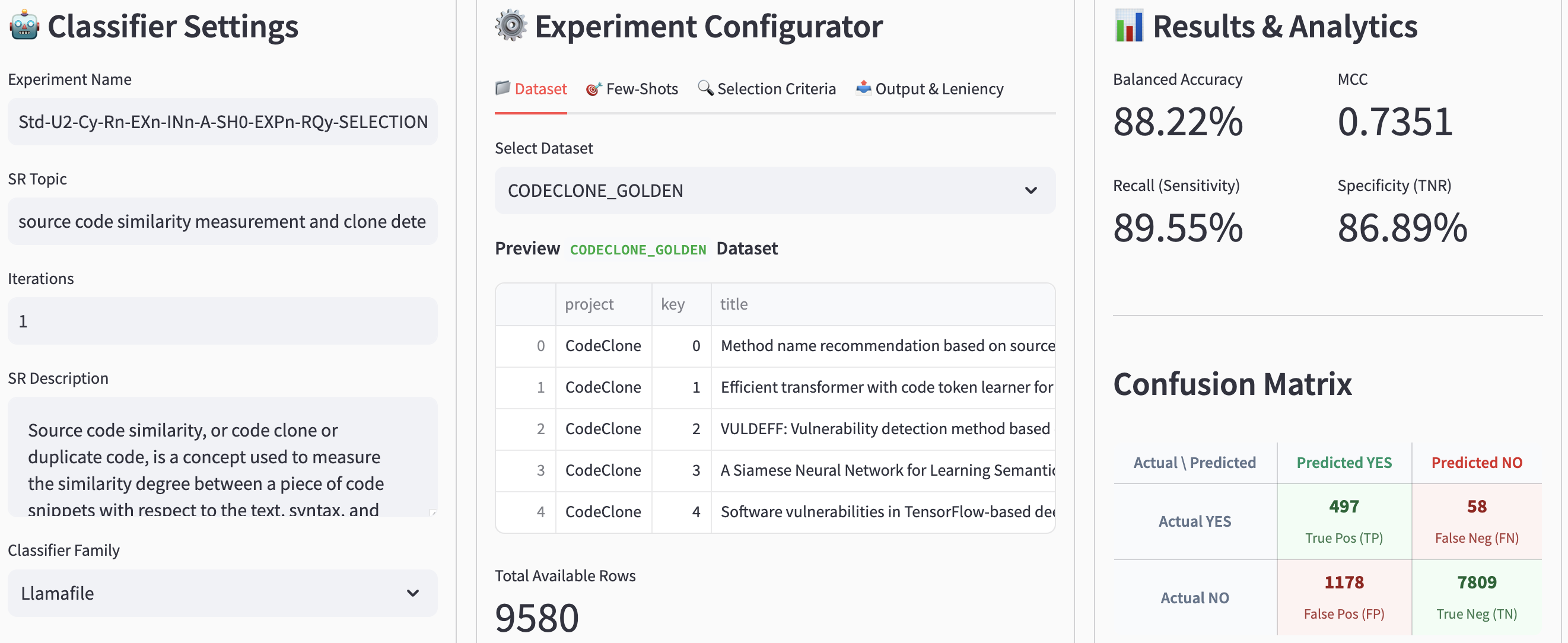}
		\caption{Configuration UI}
		\label{fig:promptslr-ui-1}
	\end{subfigure}
	\hfill
	\begin{subfigure}[t]{.75\linewidth}
		\centering
		\includegraphics[width=\linewidth]{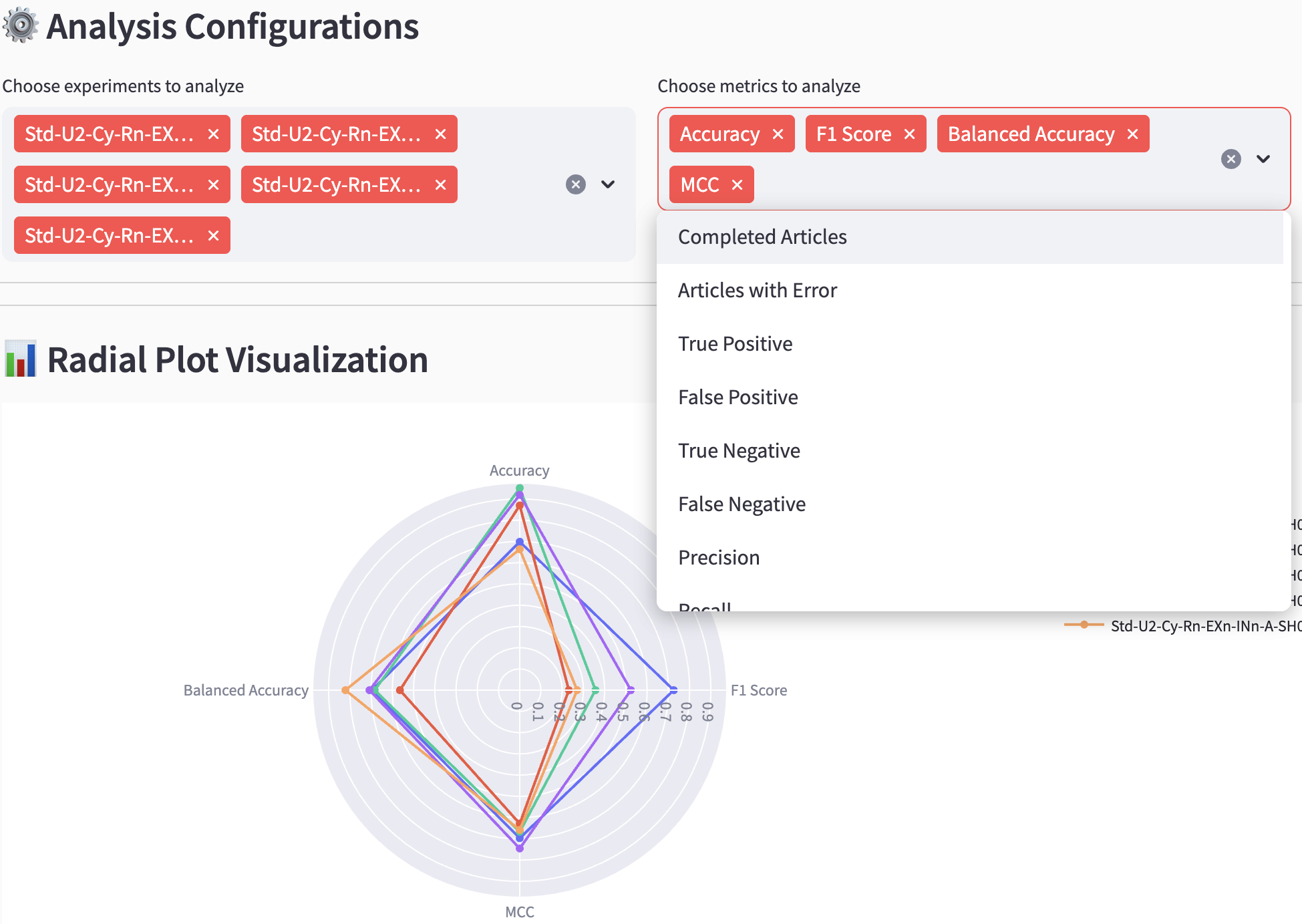}
		\caption{Analysis UI}
		\label{fig:promptslr-ui-2}
	\end{subfigure}
	\caption{PromptSR web interface: configuration (a) and result analysis (b).}
	\label{fig:promptslr-ui-merged}
\end{figure*}

These features can be toggled individually or in combination, making PromptSR directly suited for ablation studies that isolate the contribution of each feature or feature group to screening performance. Beyond prompt structure, experiments are parameterized with dataset-specific contextual information, and the system supports multiple model backends, including LLMs, traditional machine learning classifiers, and random baselines for comparison. Experiments are executed via a multithreaded scheduler, deployable locally or on HPC computing clusters, and all results, including model predictions, confidence scores, and results for all evaluation metrics, are stored in a structured database to ensure reproducibility.
The analysis module evaluates results using standard classification metrics and SR-specific measures, with MCC for comparing prompt variants and BAcc for reporting screening performance. In this context, Figure~\ref{fig:promptslr-ui-merged} shows the visualization options, including the choice of metrics, as well as tables, radial plots, box plots, scatter plots, line plots, and pie charts to support flexible result exploration.

\section{Use Case Application}\label{sec:results}

To demonstrate how SRBench and PromptSR work together in practice, we present a use case application in which we configure a prompt experiment, run it across all 32 SRs in the SRBench dataset, and analyze the results using PromptSR's built-in analysis module.

\subsection{Procedure}
The prompt used in the evaluation relies on two feature groups: the \textit{Context} component, and \textit{Selection Criteria}, which comprises only the exclusion criteria drawn directly from each dataset. No few-shot examples, chain-of-thought reasoning, or uncertainty outputs are enabled, making this a minimal but representative configuration to benchmark against SRBench. The choice of feature groups was based on the findings of Syriani~\etal~\cite{Syriani2024}, which showed that simple prompts with exclusion criteria were effective. The experiments are configured through PromptSR's web interface, where each dataset is uploaded along with its ground-truth labels and domain-specific criteria, which are entered into the corresponding prompt fields. \code{Qwen3-30B-A3B-Thinking-2507} model is selected as the evaluation model running via the \code{vllm} on a single Nvidia H100 GPU in a National High Performance Computing Center. The model was chosen for its performance and efficiency at the time of the experiment, as it performed well on the general LLM benchmark\footnote{\url{https://arena.ai/leaderboard/}, \\ \url{https://huggingface.co/Qwen/Qwen3-30B-A3B-Thinking-2507}} tasks while being smaller that can fit on a single GPU, reducing inference costs. Once configured, PromptSR's scheduler dispatches the experiment across all 32 datasets in SRBench using a multithreaded execution pipeline. Each article title and abstract is independently screened by the model using the rendered prompt, and predictions are stored alongside confidence scores, token usage, and response metadata. The full execution is logged in the database.

\subsection{Results and Analysis}

The results are analyzed using PromptSR's analysis module. The median BAcc across all datasets is 0.69, and the median MCC is 0.32, indicating that even this minimal prompt configuration produces useful screening decisions across a diverse range of software engineering topics. Table~\ref{tab:results} reports an excerpt of the screening performance considering the best and worst MCC results per SR.

Strong performance is observed on datasets with well-defined research scopes and moderate class imbalance. \code{MODELAUTOCODE4WIRELESS} achieves a  BAcc of 0.88 and an MCC of 0.75, while \code{RL4SE} reaches the highest  BAcc of 0.91 with an MCC of 0.58. \code{CODESMELLDETECT} (MCC 0.49), \code{OPINIONMINESD} (MCC 0.49), and \code{MPM4CPS} (MCC 0.48) similarly show reliable discrimination.


The worst MCC results suggest that the metrics in SRBench can reveal behavior that would likely remain hidden if only traditional metrics were used. For example, \code{LEARNSOFTCONFSPEC} achieves an apparently excellent accuracy (0.971) and F1 score (0.985). However, its MCC is 0.00, and its BAcc is below random (0.485). Similarly, \code{TESTNN} has a moderate F1 score (0.595) despite an MCC of 0.00 and a BAcc of only 0.211. These cases show that the model collapses to predicting a single class, a behavior that would not necessarily be visible when evaluating only conventional metrics. The expanded benchmark therefore provides insight into \emph{when} prompt-based screening fails, not only \emph{how well} it performs on average. These failure cases tend to involve either very small datasets or atypically high inclusion rates that the prompt, without few-shot examples or uncertainty features, cannot handle. Overall, the results support the claim that PromptSR's richer metrics allow for a more trustworthy interpretation of screening quality and help identify classifier behaviors that standard metrics alone would overlook.

Furthermore, Table~\ref{tab:results} also shows the results (accuracy and F1 score) from SERS evaluation~\cite{Huotala2025}, excluding the cases of newly added SRs. Overall, our results demonstrate competitive, and in most cases superior, performance despite relying on an open-source 30B-parameter LLM. In contrast, their study used a range of significantly larger, proprietary models, including Llama Maverick 400B, DeepSeek-R1-671B, and models from OpenAI and Anthropic (with unknown parameters). For example, for \code{MODELAUTOCODE4WIRELESS}, accuracy and F1 score from SERS are 0.56 and 0.53, while our results show an accuracy of 0.88 and an F1 score of 0.89. As discussed earlier, however, these metrics are not sufficient to fully capture screening performance and serve only as a reference point for comparison with their reported results. Note also that the SERS results in Table~\ref{tab:results} are an aggregate across all LLMs used in their study. Even under these conditions, our results from the use case show that a simple prompt configuration can achieve performance comparable to or better than the average of multiple larger models across a diverse set of SRs. This suggests that the choice of prompt design and evaluation framework, as facilitated by SRBench and PromptSR, can be as important as model size in determining screening performance.

\begin{table}[t]
    \centering
    \caption{Screening performance across the five best and worst cases, ranked by MCC.}
    \label{tab:results}
    \resizebox{\columnwidth}{!}{%
        \begin{tabular}{l|rr|rrrr}
            \hline

            \textbf{SR}
            & \multicolumn{2}{c|}{\textbf{SESR Results}}
            & \textbf{Accuracy}
            & \textbf{F1 Score}
            & \textbf{B Acc.}
            & \textbf{MCC} \\

            & \textbf{Acc.}
            & \textbf{F1}
            & & & & \\

            \hline

            \multicolumn{7}{c}{\textbf{Best Cases}} \\
            \hline

            \code{MODELAUTOCODE4WIRELESS}
            & 0.56 & 0.53 & 0.88 & \textbf{0.89} & \underline{0.88} & \textbf{0.75} \\

            \code{RL4SE}
            & -- & -- & \underline{0.88} & 0.57 & \textbf{0.91} & \underline{0.58} \\

            \code{CODESMELLDETECT}
            & 0.78 & 0.41 & \textbf{0.92} & 0.53 & 0.71 & 0.49 \\

            \code{OPINIONMINESD}
            & 0.75 & 0.50 & 0.79 & 0.57 & 0.80 & 0.49 \\

            \code{MPM4CPS}
            & 0.64 & 0.57 & 0.73 & \underline{0.74} & 0.73 & 0.47 \\

            \hline

            \multicolumn{7}{c}{\textbf{Worst Cases}} \\
            \hline

            \code{ESM\_2}
            & -- & -- & 0.59 & 0.49 & 0.53 & 0.07 \\

            \code{ARCHIML}
            & -- & -- & 0.26 & 0.03 & 0.55 & 0.03 \\

            \code{LEARNSOFTCONFSPEC}
            & 0.85 & 0.92 & 0.97 & 0.98 & 0.48 & 0.00 \\

            \code{TESTNN}
            & 0.40 & 0.57 & 0.42 & 0.59 & 0.21 & 0.00 \\

            \code{SE4MVP}
            & 0.34 & 0.42 & 0.25 & 0.32 & 0.34 & -0.25 \\

            \hline
        \end{tabular}%
    }
\end{table}


\subsection{Discussion and Limitations}\label{sec:Discussion}

The use case demonstrates how SRBench and PromptSR together enable systematic evaluation of LLM-assisted screening across a diverse collection of SR datasets. Even with a minimal prompt configuration, the results show that LLM-based screening can achieve useful performance across several SRs while simultaneously exposing important failure cases that would otherwise remain hidden under conventional evaluation metrics. This reinforces the importance of evaluating screening systems using metrics that explicitly account for class imbalance and asymmetric error costs.

A key implication of SRBench is its ability to evaluate the \emph{generalizability} of screening approaches across different software engineering domains. The benchmark dataset spans 35 ACM CCS categories and includes SRs with highly heterogeneous inclusion rates, dataset sizes, and research topics. The observed variability in performance across datasets suggests that prompt configurations and models that perform well in one review context may not generalize reliably to others. Consequently, SRBench can support future studies on investigating which prompt engineering strategies, model families, or reasoning mechanisms remain robust across diverse review settings, rather than overfitting to a small number of datasets.

The results also highlight the importance of richer evaluation metrics for SR screening. Cases such as \code{LEARNSOFTCONFSPEC} and \code{TESTNN} demonstrate that standard metrics such as Accuracy and F1 score can provide misleadingly optimistic interpretations when datasets are highly imbalanced or fully inclusive. In contrast, MCC and BAcc expose potential issues such as majority-class collapse and lack of meaningful discrimination. This suggests that future evaluations of LLM-assisted screening should prioritize imbalance-aware metrics to avoid overstating model effectiveness.

Beyond evaluating prompts and models, SRBench opens several research directions. The benchmark can support systematic ablation studies exploring the contribution of prompt features such as few-shot examples, chain-of-thought reasoning, uncertainty outputs, and leniency strategies. It also enables comparisons between proprietary and open-source LLMs, traditional machine learning classifiers, and hybrid human-in-the-loop screening workflows. Since PromptSR stores confidence scores and prediction metadata, future work could additionally investigate calibration quality, selective prediction strategies, and adaptive reviewer assistance mechanisms.

Despite these contributions, the experimental evaluation only considers a single model and a single baseline prompting strategy. While this was sufficient to demonstrate the capabilities of SRBench and PromptSR, it does not constitute an exhaustive assessment of LLM-assisted screening performance. Different models, prompting techniques, sampling parameters, or reasoning strategies may produce substantially different results. The goal of the use case is therefore to provide a proof of concept for the framework rather than a definitive comparison of screening methods.

Benchmark construction demands substantial manual effort and is inherently error-prone. As evidence, beyond the 14 newly added SRs, we applied the same curation process described in Section~\ref{sec:slrbench} to five existing SESR entries and found errors requiring correction. This analysis revealed several inconsistencies in the inclusion and exclusion labels across the original replication package published alongside the SRs analyzed. This explains a few edge cases, such as the \code{TESTNN} dataset, which has a zero number of exclusions. Since our data-gathering process depends not only on the availability of replication packages but also on author responses, validation required extensive curation and correction of inconsistencies across sources. Extending this approach to other scientific domains would therefore require considerable community effort and domain expertise.
Nevertheless, using the same methodology to add new SRs to our dataset would extend the benchmark's generalizability beyond software engineering. PromptSR, on the other hand, can be easily extended to new datasets from different domains like medicine~\cite{wang2025accelerating} by merely changing the prompt template. This flexibility allows researchers to use our benchmarking framework as a standardized evaluation platform, which would benefit proposals focused on prompt engineering methodologies~\cite{Syriani2024,Trad2025}.

Finally, the benchmark remains limited to title and abstract screening decisions derived from completed SRs. It does not capture downstream review stages such as full-text screening, data extraction, or quality assessment. Future extensions of SRBench could incorporate these aspects to evaluate whether LLMs can support later phases of evidence synthesis beyond initial screening. Similarly, PromptSR could be extended with workflows for extracting structured evidence, identifying missing information, supporting reviewer disagreement resolution, and generating traceable rationales for inclusion or exclusion decisions. Such extensions would enable the benchmark to support end-to-end evaluation of AI-assisted SR pipelines rather than only title and abstract classification.

\section{Conclusion}\label{sec:Conclusion}
In this paper, we presented a benchmarking framework for evaluating screening automation in software engineering SRs. Our benchmark relies on imbalance-aware evaluation metrics to provide a more trustworthy interpretation of screening performance and includes a dataset of $45\,064$ labeled entries from 32 SRs. We also introduced PromptSR, a tool for prompt experimentation, execution, and analysis. Together, they support reproducible and controlled evaluation of screening approaches across diverse SR datasets. In future work, we will focus on expanding SRBench with additional domains and review types, evaluating more LLMs and different prompting strategies, and improving automation in the screening pipeline, particularly for dataset curation and error handling in large-scale experiments. We also plan to make the benchmark publicly available through an online platform to encourage community contributions and facilitate ongoing research on LLM-assisted screening in evidence synthesis.

\section*{Data Availability}

A replication package, including the dataset and source code, is available at~\cite{10.5281/ZENODO.20434525}.

\bibliographystyle{ACM-Reference-Format}
\bibliography{references}

\end{document}